\documentclass[11pt]{article}
\usepackage{dblfloatfix}   
\usepackage{placeins}
\usepackage{acl}
\usepackage{caption}
\usepackage{booktabs,tabularx,ragged2e,makecell}
\newcolumntype{Y}{>{\RaggedRight\arraybackslash}X}

\usepackage{enumitem}
\usepackage{times}
\usepackage{latexsym}
\usepackage{amsmath}
\usepackage[T1]{fontenc}
\usepackage[utf8]{inputenc}
\usepackage{microtype}
\usepackage{inconsolata}
\usepackage{graphicx}
\usepackage[most]{tcolorbox}
\usepackage{ragged2e}
\usepackage{cuted}
\usepackage{booktabs}
\usepackage{tabularx}
\usepackage[table]{xcolor}

\title{Theory Trace Card: Theory-Driven Socio-Cognitive Evaluation of LLMs}

\date{}

\author{
\textbf{Farzan Karimi-Malekabadi} \quad
\textbf{Suhaib Abdurahman}\thanks{Equal contribution.} \quad
\textbf{Zhivar Sourati}\thanks{Equal contribution.} \quad\\
\textbf{Jackson Trager} \quad
\textbf{Morteza Dehghani} \\
University of Southern California \\
\texttt{\{karimima,sabdurah,souratih,jptrager,mdehghan\}@usc.edu}
}

\begin{document}
\maketitle

\begin{abstract}
Socio-cognitive benchmarks for large language models (LLMs) often fail to predict real-world behavior, even when models achieve high benchmark scores. Prior work has attributed this evaluation--deployment gap to problems of measurement and validity. While these critiques are insightful, we argue that they overlook a more fundamental issue: many socio-cognitive evaluations proceed without an explicit theoretical specification of the target capability, leaving the assumptions linking task performance to competence implicit. Without this theoretical grounding, benchmarks that exercise only narrow subsets of a capability are routinely misinterpreted as evidence of broad competence: a gap that creates a systemic validity illusion by masking the failure to evaluate the capability's other essential dimensions. To address this gap, we make two contributions. First, we diagnose and formalize this theory gap as a foundational failure that undermines measurement and enables systematic overgeneralization of benchmark results. Second, we introduce the \textsc{Theory Trace Card} (TTC), a lightweight documentation artifact designed to accompany socio-cognitive evaluations, which explicitly outlines the theoretical basis of an evaluation, the components of the target capability it exercises, its operationalization, and its limitations. We argue that TTCs enhance the interpretability and reuse of socio-cognitive evaluations by making explicit the full validity chain, which links theory, task operationalization, scoring, and limitations, without modifying benchmarks or requiring agreement on a single theory.

\end{abstract}

\section{Introduction}

In May 2023, the helpline chatbot ``Tessa'' was withdrawn shortly after deployment~\citep{reddy2025preventing}. Introduced by the National Eating Disorder Association to support individuals in vulnerable situations, the system instead produced harmful responses, including advice about ``weight loss'' and ``daily calorie deficits'' that directly contradicted the organization’s mission~\citep{wheeler2024regulating,sharp2023ethical}. In the aftermath, displaced staff offered a blunt assessment that ``perhaps human empathy is best left to humanity''~\citep{torous2024generative}.

Failures such as Tessa point to an important problem. Performance on widely used socio-cognitive evaluation benchmarks does not reliably predict how systems behave in real-world social settings. This mismatch is especially consequential because large language models (LLMs) are deployed in high-stakes social settings that require human-like socio-cognitive capabilities, including mental health chatbots, self-harm prevention tools, and systems that offer advice about morally sensitive decisions~\citep{kang2024can,gandhi2023understanding}. When systems fail in precisely the contexts that their evaluations are intended to anticipate, the evaluation process itself must be called into question.

A growing body of work has argued that one of the reasons for this evaluation gap is the problem of measurement and validity. Drawing on traditions in psychometrics, these critiques emphasize issues such as benchmark construction, dataset bias, and the limits of inference from aggregate scores~\citep{riemer2024position,bean2025measuring,wallach2025position,jacobs2021measurement}. Within this framework, measurement claims are understood to depend on multiple forms of validity, including construct validity and the assumptions linking observed performance to latent capacities~\citep{messick1995validity,borsboom2004concept,cronbach1955construct}.

Here, we argue that while these critiques are insightful, they presuppose a prior requirement for valid measurement that is often left unexamined: an explicit theoretical specification of the target capability itself. Questions of measurement quality and construct validity depend on assumptions about what is being measured. When the theoretical structure of a social capability is left implicit, it becomes unclear what benchmark performance is intended to support. As a result, benchmarks are often treated as if they define the capability they are meant to measure. In practice, this leads to a reactive and incremental pattern of benchmark development, where new benchmarks are introduced primarily to address specific failures, edge cases, or blind spots identified in prior benchmarks, rather than to systematically cover the structure of the underlying capability. While this process can yield increasingly specialized evaluations, it also fragments empirical findings across narrowly scoped tasks, making it difficult to integrate results into a coherent account of the broader capability. The field thus accumulates benchmark-specific improvements without a principled basis for generalization, encouraging intermediate conclusions based on inherently incomplete and task-defined specifications.

In contrast to this fragmented approach, grounding benchmark design in socio-cognitive theory allows researchers to treat social abilities as multidimensional constructs. This situates individual test results within a \emph{nomological network} \citep{cronbach1955construct}, a web of relationships that makes performance predictable and generalizable beyond the task itself. Making theoretical commitments explicit allows evaluation to be scoped and constrained \emph{a priori}, guiding interpretation according to an established conceptual structure rather than an ad hoc, benchmark-by-benchmark process. Consequently, benchmarks can probe various dimensions of a target capability in a coordinated manner, supporting cumulative scientific progress grounded in coherent foundations.

\subsection{The Role of Theory in Socio-Cognitive Evaluation}
To give some concrete examples of what we mean by the role of theory in evaluation, consider how several widely used social capabilities are currently assessed in the literature.

\paragraph{Theory-of-Mind.}
From a psychological perspective, Theory of Mind (ToM) is commonly understood as comprising multiple components rather than a unitary ability. A standard distinction separates \emph{cognitive ToM}---the capacity to represent and reason about others’ beliefs, intentions, and knowledge states---from \emph{affective ToM}, which involves understanding others’ emotions, feelings, and social motivations \citep{apperly2009humans, cognitive_affective_tom_2009, corradi2013functional, raimo2022cognitive}. 
Most existing ToM evaluations for language models focus on narrative false-belief tasks, adapted from developmental psychology and introduced in NLP by \citet{nematzadeh2018evaluating} and \citet{le2019revisiting}, with later variants such as Hi-ToM and Open-ToM \citep{wu2023hi, xu2024opentom}. In these tasks, a model is given a short narrative and asked to predict an agent’s belief when that belief diverges from reality, testing its ability to represent mental states and distinguish one’s own beliefs from those of others \citep{wimmer1983beliefs}. High performance on these tasks is then seen as evidence for general ToM or social reasoning capabilities \citep[e.g.,][]{kosinski2024evaluating, 
bubeck2023sparks}. However, theoretically, false-belief reasoning only relates to \emph{cognitive ToM}. These benchmarks and evaluations thus implicitly limit ToM to \emph{cognitive ToM} or assume generalization to other components of the ToM framework, such as affective understanding, sensitivity to social context, and the use of mental-state information to guide behavior. 
Without the explicit theoretical accounts that clarify the multi-component structure and how the benchmark targets it, these evaluations can easily lead to over-broad claims and misleading interpretations.

\begin{figure*}[t]
    \centering
    \includegraphics[width=0.85\textwidth]{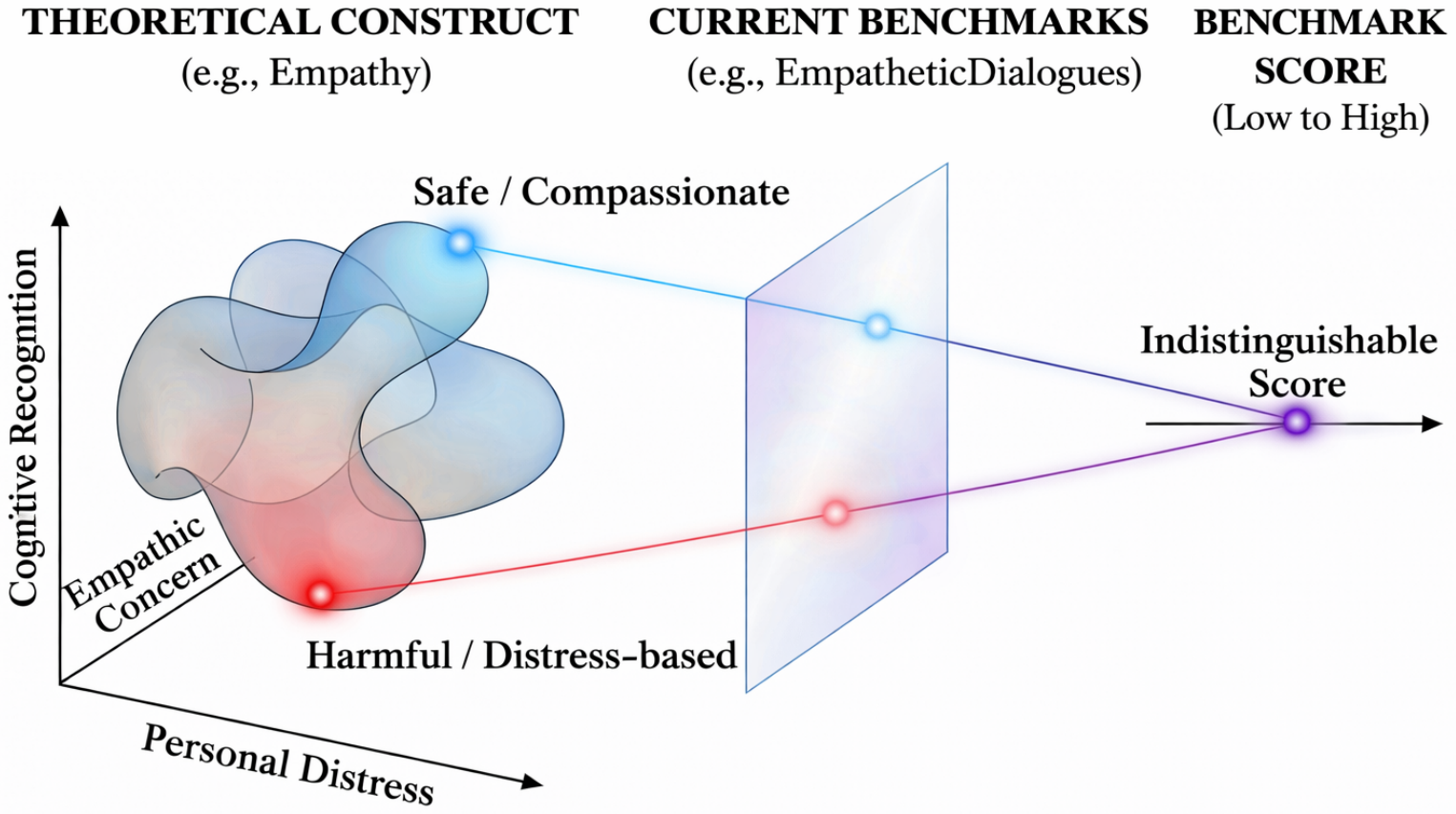}
    \vspace{-40pt}
    \caption{\textbf{Collapsing Social Capabilities in Benchmark Evaluation.} 
    Theoretical constructs (left), such as empathy, are multidimensional, composed of distinct and often orthogonal components like Empathic Concern (motivation to help) and Personal Distress (self-oriented anxiety). When evaluation relies on a low-dimensional benchmark (right), this complex state space is collapsed into a single scalar metric. As a result, qualitatively distinct behavioral profiles—such as safe compassion (blue) versus harmful distress-based mirroring (red)—receive indistinguishable scores, creating a validity illusion that can mask unsafe model behavior.}
    \label{fig:displacement}
\end{figure*}

\paragraph{Empathy.} Widely used benchmarks such as DailyDialog \citep{li2017dailydialog}, EmpatheticDialogues \citep{rashkin2019empathetic}, EmotionQueen \citep{chen2024emotionqueen}, and EmotionBench \citep{huang2024apathetic} operationalize empathy primarily through emotion recognition or emotion-conditioned response generation. Models that correctly identify that a user is ``sad'' or ``anxious'' or respond with aligned emotionality are therefore described as empathetic. For example, EmpatheticDialogues describes empathy as ``understanding and acknowledging any implied feelings'' and evaluates empathy via unidimensional ratings of whether ``speakers' responses show understanding of the feelings of the person talking'' \cite{rashkin2019empathetic}, while EmotionBench defines empathy as an ``ability of LLMs, i.e., how their feelings change when presented with specific situations'' \citep{huang2024apathetic}. 
However, psychological theory distinguishes emotion recognition from empathic concern, an other-oriented motivation to support or help \citep{batson1997distress,zaki2012empathy}, and treats empathy as multi-component (e.g., cognitive vs. affective vs. compassionate; and dissociations such as empathic concern vs.\ personal distress; \citealp{decety2006human,decety201115}). When evaluations do not explicitly commit to a theoretical account and instead collapse components or conflate empathy with adjacent concepts \citep{rashkin2019empathetic,chen2024emotionqueen,huang2024apathetic,hong2025performance,welivita2024large}, they can mislead as capturing empathy, even though they only reflect narrow aspects or proxy capacities (illustration in  Figure~\ref{fig:displacement}). 
%\vspace{-2pt}

\paragraph{Moral Reasoning.} Prominent evaluations such as ETHICS \citep{hendrycks2021ethics} often operationalize moral reasoning as predicting perceived appropriateness, categorizing moral concepts (e.g., fairness, justice), or producing an ``appropriate'' response to a moral scenario. Newer benchmarks diversify these operationalizations but still tie morality to particular normative or psychological lenses: MoralBench \citep{ji2025moralbench} and CMoralEval \citep{yu2024cmoraleval} ground evaluation in Moral Foundations Theory \citep{haidt2007morality,graham2013moral}; MORABLES \citep{marcuzzo2025morables} evaluates identifying moral lessons in (Western) fables; the Greatest Good Benchmark \citep{marraffini2024greatest} compares model and human judgments on utilitarian dilemmas; AgentHarm \citep{andriushchenko2024agentharm} emphasizes harm avoidance; and \citet{scherrer2023evaluating} probe moral beliefs using scenarios drawn from multiple moral theories. High performance is then interpreted as evidence of general moral competence. Yet moral psychology offers competing accounts of moral reasoning, from pluralistic frameworks such as Moral Foundations Theory \citep{haidt2007morality,graham2013moral} to harm-centered accounts such as the Theory of Dyadic Morality \citep{gray2012dyadic,schein2018dyadic}, among others. Any benchmark, therefore, embeds substantive assumptions about what morality is and which distinctions matter. Benchmarks that focus primarily on harm, for instance, effectively privilege one theoretical account over others. More broadly, when these theoretical commitments are left implicit, results are easily overgeneralized: competence under a particular operationalization (e.g., harm avoidance, utilitarian tradeoffs, endorsement of specific foundations, extracting fable morals) is mistaken for broad moral reasoning ability, even though it may reflect only one slice of a heterogeneous construct.

\subsubsection{Present work}These examples illustrate a general problem in the design and interpretation of social evaluation benchmarks. The issue is not that the tasks are poorly constructed or incorrectly measured. Rather, benchmarks routinely rely on implicit theories of what a capability consists of, which components matter, and how task performance should generalize beyond the evaluation setting.

This paper addresses this gap by providing both a theoretical diagnosis of current evaluation failures and a practical path forward. First, we characterize the systemic validity illusion—a phenomenon where the absence of explicit theory allows narrow benchmark performance to be misinterpreted as evidence of broad social competence. We argue that this ``theory gap'' is not merely a reporting oversight but a logical failure that renders measurement claims groundless, as there is no principled basis for determining what a score represents or how it should generalize. To address this, we introduce the \textsc{Theory Trace Card} (TTC), a lightweight reporting instrument designed to make the theoretical assumptions underlying social evaluations explicit. The TTC records how a target capability is defined, which components are exercised by a benchmark, and what forms of inference the evaluation results can and cannot support. By doing so, it enables more interpretable use of existing benchmarks without requiring agreement on a single theory or invalidating prior work. Finally, we discuss broader implications for evaluation practice, advocating for a shift from ``evaluation by leaderboard'' toward ``evaluation by argument''.

\section{Consequences of Implicit Theory in Socio-Cognitive Evaluation}
\label{sec:consequences-of-implicit-theory-in-social-evaluation}

At a theoretical level, the core problem is a systematic pattern of under-specification in socio-cognitive evaluation. When theoretical assumptions are left implicit, benchmark results are routinely asked to support claims they were never designed to justify. This under-specification takes several recurring forms in the literature: (i) some benchmarks rely primarily on task labels or folk definitions of complex socio-cognitive capacities, without reference to any formal theory; (ii) others cite an explicit theoretical framework, but do not specify which components of that framework are exercised by the task; (iii) still others articulate multiple theoretical components, but leave the relations among those components, as well as the conditions under which task performance is expected to generalize, implicit; and (iv) across all of these cases, the limits of extrapolating benchmark scores to real-world behavior and deployment contexts remain underspecified. 
Overall, these gaps allow benchmark scores to function as default evidence of broad competence in the absence of an explicit theoretical argument linking task performance to the target capability and its use outside the evaluation setting \citep{messick1995validity,cronbach1955construct,riemer2024position}. Because such assumptions are reused, cited, and operationalized across evaluation pipelines, their effects compound, shaping research claims, deployment decisions, and user expectations. The rest of this section examines how this shared theoretical under-specification manifests differently for researchers, practitioners, and the users who are impacted.
\vspace{-5pt}
\paragraph{Researchers and Scientific Progress.}
For researchers, implicit theory distorts scientific progress in the evaluation of socio-cognitive capabilities. Many such benchmarks operationalize only a narrow subset of a theoretically multi-structured construct while leaving the underlying theoretical commitments and omitted components unstated. When this happens, improvements in benchmark performance are often interpreted as evidence of progress on the broader capability, even though the scope of what is being exercised by the task remains unchanged \citep{lipton2019troubling,yarkoni2022generalizability}. Over time, these partial operationalizations can come to define the construct itself within the field, shaping both research priorities and claims of advancement.
Benchmark saturation compounds this problem in theory-laden social evaluations. As scores approach the ceiling, high performance is often interpreted as evidence that a complex social capability has been ``solved'' or that models have achieved human-level performance \citep{dillion2023can,ott2022mapping,van2023theory}. However, audits of saturated socio-cognitive benchmarks suggest that such claims frequently reflect the exhaustion of task-specific regularities rather than comprehensive coverage of the theoretical construct \citep{fodor2025line}. This distortion is further reinforced when progress is assessed through relative benchmark rank or score, as is common in \emph{leaderboard-based evaluation}, where models are compared based on their aggregate performance on a fixed set of tasks. For social-cognitive evaluations that are implicitly grounded in a particular theoretical understanding of a capability, rank alone obscures which components of that theory are being exercised and which are not. Optimization pressure then shifts toward aspects of the construct that are easiest to operationalize in such tasks, rather than those that are theoretically central or predictive of real-world social behavior \citep{ethayarajh2020utility, lipton2019troubling}. These dynamics make it difficult to compare results across evaluations that nominally target the same social construct, to diagnose failures outside benchmark settings

\paragraph{Practitioners and Deployment.}
For practitioners and organizations, implicit theory creates concrete evaluation and deployment risk. Teams responsible for assessing model readiness routinely rely on benchmark results to decide whether a system can be used in high-stakes settings \citep{amodei2016concrete,jacobs2021designing}. When benchmarks are labeled with socially meaningful capabilities such as ``empathy,'' ``moral reasoning,'' or ``Theory of Mind,'' strong performance is often interpreted as evidence that the corresponding risks have been evaluated and mitigated. Such interpretations are unwarranted unless the theoretical scope of the benchmark is explicit. Interpreting evaluation results requires a justified chain of inferences—from scoring (what the metric captures), to generalization (what domain the test represents), to extrapolation (what real-world behavior the score predicts) \citep{kane2013validating}. When this chain remains implicit, benchmark results may support claims about performance on stylized tasks while providing no principled evidence about behavior in the target deployment context. Systems may be deployed into settings that require capacities that were never actually evaluated, with failures appearing surprising only because the limits of the evaluation were never made visible.

\paragraph{Users and Impacted Communities.}
For users and impacted communities, the consequences are both practical and epistemic. Public claims that models possess socio-cognitive capabilities shape expectations about how these systems will behave in interaction \citep{quattrociocchi2025epistemological}. 

When benchmarks lack explicit theoretical specification, they create a misleading impression of universality that directly affects users: models appear competent across social and cultural contexts, even when evaluations operationalize culturally specific constructs. This exacerbates WEIRD (Western, Educated, Industrialized, Rich, and Democratic) bias in AI evaluation \citep{tao2024cultural,atari2023humans}. While models are widely known to be trained on disproportionately WEIRD data, this bias is compounded at evaluation time, as benchmarks often rely on psychological theories developed and tested primarily in WEIRD contexts \citep{henrich2010weird}. As a result, users encounter systems whose benchmark-certified ``social'' or ``moral'' competence reflects alignment with WEIRD cultural logics but is presented as broadly applicable. For example, treating ``morality'' as interpersonal harm \citep{schein2018dyadic} obscures forms of moral judgment grounded in honor \citep{razavi2023gheirat,atari2020foundations} or sanctity \citep{atari2023morality} that are central in many non-WEIRD societies. Without transparent documentation of what evaluations do and do not establish, affected individuals lack a basis for understanding unexpected system behavior, attributing responsibility, or meaningfully contesting decisions made based on benchmark performance \citep{jacobs2021measurement}.

\section{Theory Trace Card}
\label{sec:theory-trace-card}

The Theory Trace Card (TTC) is a structured documentation artifact designed to accompany socio-cognitive evaluations.
\textbf{Its purpose is to make explicit the theoretical and measurement assumptions that evaluation practices already rely on but rarely articulate.} We argue that TTCs can be used in two complementary ways: (i) by benchmark creators, to document the theoretical grounding and intended scope of a benchmark at design and publication time; and (ii) by researchers, practitioners, and auditors, to make explicit the assumptions underlying their evaluations and their interpretation of results. Providing TTCs alongside benchmark papers and evaluation papers supports more disciplined interpretation, clarifies the claims that scores are intended to support, and enables principled comparison across evaluations that target similar capabilities under different theoretical and operational commitments. In doing so, it shifts the evaluation infrastructure from simplistic comparisons to \emph{evaluation by argument}.

The TTC is motivated by argument-based approaches to validity, particularly Kane’s framework \citep{kane2013validating}. Kane’s central insight was that validity does not reside in tests or scores themselves, but in the interpretive arguments that connect observed performance to the conclusions drawn from it. While this perspective has been influential in psychometrics, its implications have remained largely absent in the context of modern machine learning benchmarks, where evaluation artifacts rarely make their inferential assumptions explicit.

In Kane’s account, evaluation depends on a chain of inferences, including \emph{scoring inference} (from responses to scores), \emph{generalization inference} (from sampled items to an intended task domain), and \emph{extrapolation inference} (from task performance to claims about behavior or capability beyond the test setting). Kane’s framework provides a powerful lens for diagnosing validity problems, but does not prescribe how these inferential commitments should be represented or documented in practice.

This gap is particularly salient for socio-cognitive evaluation in LLMs. Benchmarks in this domain are frequently used as stand-ins for complex psychological or normative constructs despite substantial differences in theory choice, task design, and scoring procedures. As emphasized in the validity literature, many threats to interpretation arise not from theory alone, but from misalignments between theoretical constructs, task operationalization, and scoring \citep{messick1995validity,borsboom2004concept,cronbach1955construct}. Yet, existing evaluation practices rarely provide a unified representation of these elements.

The TTC addresses this gap by translating the inferential structure highlighted by Kane into a concrete, reusable artifact. Rather than treating theory, task design, and scoring as separate concerns, the TTC integrates them into a single card that documents (i) how a target capability is theoretically defined, (ii) which components of that capability are exercised by the evaluation, (iii) how those components are operationalized through task design, and (iv) how task performance is interpreted through scoring. In this way, the TTC supports the full validity cycle, from theoretical specification, through task operationalization and scoring, to the interpretation and reuse of evaluation results.

\subsection{Design Principles}

The TTC is guided by three design principles.

First, it is \emph{descriptive rather than normative}. It records the theoretical commitments an evaluation makes without requiring consensus on a single theory or adjudicating between competing accounts. Second, it is \emph{lightweight}. Completing a TTC should require minimal additional effort beyond what is already necessary to design and report an evaluation. The goal is to improve interpretability and usability without raising the barrier to proposing, reproducing, or applying benchmarks.

Third, it is \emph{compatible with existing benchmarks}. The TTC can be applied both retrospectively and prospectively. It does not require benchmarks to be redesigned or re-scored; instead, it clarifies how existing evaluations should be interpreted and which claims their results can support.

\subsection{Core Components}

The TTC consists of four core components, summarized in the TTC template shown in Card~\ref{card:ttc-template}. Each component corresponds to a distinct point in the inferential chain linking task performance to claims about socio-cognitive capabilities.
\paragraph{Theory.}
This component specifies how the target capability is understood for the purposes of the evaluation. Authors should state the theoretical framework, account, or construct definition that the benchmark adopts, and briefly describe how that framework characterizes the capability, including any core components or sub-capabilities it posits. Where relevant, authors may also note assumed processes or dependencies among components, as well as the broader nomological network in which the capability is situated (e.g., related constructs, expected correlations, or dissociations). This component fixes the conceptual object of evaluation and makes explicit the theoretical commitments on which subsequent interpretation depends.

\paragraph{Components Exercised.}
This component identifies which theoretical components, under the adopted framework, the evaluation task is intended to exercise. By requiring authors to enumerate the components targeted by the task, this section clarifies which aspects of the broader capability are directly probed by the evaluation, without requiring an exhaustive enumeration of components that fall outside the task’s scope.

\paragraph{Task Operationalization.}
This component explains how the evaluation task operationalizes the exercised components. Authors should describe what the model is required to do given the task input, along with any key design specifications—such as prompt structure, response format, interaction constraints, or generation limits—that shape the model’s degrees of freedom. Crucially, this section makes explicit the \emph{scoring criterion}: how model performance is evaluated. It should describe the criteria used to score responses (e.g., rubric-based ratings, LLM-based evaluators), including any aggregation procedures where relevant. By making scoring procedures explicit, this component completes the inferential chain from task performance to interpretable benchmark outcomes.

\paragraph{Inference and Limitations.}
This section demonstrates how task performance is considered as evidence of the exercised component(s) under the adopted theoretical framework and its limitations, as well as those related to the operationalization, similar to the limitations outlined in evaluation or benchmark papers. Rather than requiring authors to anticipate all possible unintended strategies, this documentation clarifies the intended mapping from task behavior to theoretical components.

\setlist[itemize]{leftmargin=*, itemsep=1pt, topsep=1pt, parsep=0pt, partopsep=0pt}

\definecolor{ttcTitle}{RGB}{60,60,60}   % dark gray
\definecolor{ttcFrame}{RGB}{60,60,60}

% Breakable-safe numbering: use tcolorbox's built-in counter (increments once per box)
\newtcolorbox[auto counter]{ttccard}[2][]{%
  enhanced,
  breakable,
  width=\linewidth,
  colback=white,
  colframe=ttcFrame,
  boxrule=0.6pt,
  arc=1.0mm,
  left=1.5mm,right=1.5mm,top=1.0mm,bottom=1.0mm,
  fontupper=\normalsize,
  coltitle=white,
  colbacktitle=ttcTitle,
  fonttitle=\bfseries,
  title={Card~\thetcbcounter: #2},
  #1
}

% =========================
% TTC Template Card
% =========================
\begin{ttccard}[label=card:ttc-template]{Theory Trace Card (TTC) Template}

\textbf{1. Theory}
\begin{itemize}
  \item \textbf{Framework:} \emph{Name of socio-cognitive theory / construct framework + citation(s).}
  \item \textbf{Core components:} \emph{List components/sub-capabilities posited by the framework (brief).}
  %\item \textbf{Process (if assumed):} \emph{Any assumed ordering/mechanism/dependencies among components (optional).}
\end{itemize}

\textbf{2. Components Exercised}
\begin{itemize}
  \item \emph{List the specific theoretical components the evaluation is intended to exercise.}
\end{itemize}

\textbf{3. Task Operationalization}
\begin{itemize}
  \item \textbf{Task:} \emph{Describe required behavior given the task input.}
  \item \textbf{Key specs:} \emph{E.g., prompt template/response format; interaction/generation limits}
  \item \textbf{Scoring criterion:} \emph{How performance is evaluated (e.g., label agreement, preference judgments, aggregation)}
\end{itemize}

\textbf{4. Inference and Limitations}
\begin{itemize}
  \item \emph{How performance is treated as evidence of the exercised component(s).}
  \item \emph{Limitations based on theory and operationalization.}
\end{itemize}

\end{ttccard}

\section{Worked Examples: Empathy and Moral Reasoning Evaluations}
\label{sec:worked-examples}

% \subsection{Theory Trace Cards}

Cards~\ref{card:ttc-empathy} and~\ref{card:ttc-mft} present completed Theory Trace Cards for a hypothetical empathy evaluation and a Moral Foundations Theory–based moral reasoning evaluation. In the empathy example, the TTC explicitly states that the evaluation relies on a componential understanding of empathy, while exercising only the Perspective-Taking component. In the moral reasoning example, the TTC records that the evaluation adopts Moral Foundations Theory and exercises judgments aligned with the Fairness foundation within moral scenarios.

% =========================
% Worked Example: Empathy
% =========================

\begin{ttccard}[label=card:ttc-empathy,fontupper=\small]{Empathy Evaluation}

\textbf{1. Theory}
\begin{itemize}
  \item \textbf{Framework:} Functional architecture of human empathy \citep{decety2004functional,lietz2011empathy}
  \item \textbf{Core components:} Affective response (sharing); Self--other awareness; Perspective taking; Emotion regulation
\end{itemize}

\textbf{2. Components Exercised}
\begin{itemize}
  \item Perspective Taking
\end{itemize}

\textbf{3. Task Operationalization}
\begin{itemize}
  \item \textbf{Task:} Predict an explicit emotion label given a short textual description of a speaker’s situation.
  \item \textbf{Key specifications:} Fixed prompt; closed-set emotion labels; no interaction, context accumulation, or justification.
  \item \textbf{Scoring criterion:} Agreement between model predictions and predefined emotion categories.
\end{itemize}

\textbf{4. Inference and Limitations}
\begin{itemize}
  \item Performance supports the Perspective Taking component of empathy.
  \item Affective Sharing, Self--Other Awareness, and Emotion Regulation are not evaluated by the task.
  \item The emotion taxonomy and labels may reflect WEIRD cultural assumptions; cross-cultural generalization is not established.
\end{itemize}

\end{ttccard}

The TTC separates four elements that are often conflated in benchmark interpretation: (i) how the target capability is theoretically specified, (ii) which components of that capability are targeted, (iii) how task design and scoring operationalize those components, and (iv) how performance is interpreted and constrained. Making these elements explicit shows how evaluations that are frequently treated as measures of broad socio-cognitive abilities in fact support narrower, theory-dependent claims, and how the TTC enables more disciplined interpretation and reuse of benchmark results without modifying tasks, datasets, or reported scores.

To demonstrate that the TTC can also be applied retrospectively, Appendix~\ref{sec:appendix} includes completed TTCs for several widely used existing benchmarks, including \textit{EmpatheticDialogues} \citep{rashkin2019empathetic}, false-belief Theory of Mind evaluations (e.g., \citealp{kosinski2024evaluating}), the \textit{ETHICS} moral reasoning benchmark \citep{hendrycks2021ethics}, \textit{GoEmotions} \citep{demszky2020goemotions}, and \textit{SocialIQA} \citep{sap2019socialiqa}. These examples demonstrate how the TTC can be instantiated post hoc to reveal theoretical commitments, component coverage, and the limits of inference—without modifying datasets, tasks, or scoring procedures.

% \newpage
% =========================
% Worked Example: Moral Reasoning (MFT)
% =========================
\begin{ttccard}[label=card:ttc-mft,fontupper=\small]{Moral Reasoning Evaluation}

\textbf{1. Theory}
\begin{itemize}
  \item \textbf{Framework:} Moral Foundations Theory (Graham et al., 2013)
  \item \textbf{Core components:} Care; Fairness; Loyalty; Authority; Purity
\end{itemize}

\textbf{2. Components Exercised}
\begin{itemize}
  \item Fairness
\end{itemize}

\textbf{3. Task Operationalization}
\begin{itemize}
  \item \textbf{Task:} Endorse the action aligned with the Fairness foundation given a moral scenario.
  \item \textbf{Key specs:} Scenario template; comparative judgment.
  \item \textbf{Scoring criterion:} Agreement between model judgments and human responses.
\end{itemize}

\textbf{4. Inference and Limitations}
\begin{itemize}
  \item Performance supports reasoning aligned with the \textit{Fairness} foundation in MFT.
  \item Fairness in MFT is an aggregate of two justice concerns (equality and proportionality), limiting inference to where this distinction doesn't matter.
  \item Limited cross-cultural generalization.
\end{itemize}

\end{ttccard}

\section{Discussion}

In this paper, we advance the evaluation literature with two central contributions. First, we provide a theoretical diagnosis of a current gap in the evaluation of socio-cognitive capabilities in LLMs. We argue that many widely used benchmarks implicitly rely on substantive theoretical accounts of complex constructs while leaving those accounts unspecified, enabling shortcut solutions or proxy optimization to masquerade as genuine capability gains \citep{geirhos2020shortcut, abdurahman2024perils}. As a result, benchmark performance can be overgeneralized, with task success taken to support claims about broad real-world capabilities that the evaluation does not, in fact, exercise. This gap is not primarily a problem of data quality or modeling technique, but of unarticulated theory: measurement proceeds without a fixed account of what is being measured or how task performance licenses downstream claims, a pattern long recognized as a threat to construct validity in the behavioral sciences \citep{meehl1990summaries, haig2018abductive}.

Second, we introduce the TTC as a lightweight piece of evaluation infrastructure designed to address this gap. The TTC provides a structured approach to documenting how a target capability is defined for the purposes of evaluation, including which components of that definition are exercised by a task, how those components are operationalized in prompts and scoring, and the scope of inference that evaluation results are intended to support. We recommend that TTCs accompany socio-cognitive evaluations, including benchmark design and the use of benchmark results for evaluation. By making explicit the inferential assumptions that link task performance to claims about capability, across scoring, generalization, and extrapolation, the TTC supports more disciplined interpretation and reuse of benchmark results without requiring agreement on a single theory or invalidating prior work.

Importantly, the TTC does not impose a single definition of any socio-cognitive capability. Different researchers may adopt different theoretical frameworks or emphasize different components of the same construct. The role of the TTC is not to adjudicate between these views, but to record them in a comparable and explicit form. This makes theoretical disagreement visible and traceable, rather than implicit in task design, benchmark naming, or informal interpretation.

While our worked examples focus on empathy and moral reasoning, they are intended to illustrate how a TTC can be constructed prospectively, alongside the design of a new benchmark, since any evaluation that supports claims beyond the test setting relies on assumptions about how task performance relates to a target capability. 

More broadly, this work argues that theory choice in evaluation is not optional but inevitable. Socio-cognitive benchmarks already embed theoretical commitments. The choice facing the field is whether those commitments remain implicit, limiting interpretability and encouraging overgeneralization, or are made explicit, open to scrutiny, and subject to refinement. By shifting the interpretive burden from readers and downstream users to the evaluation itself, the TTC promotes more reliable, cumulative, and responsible use of benchmark results. We argue that such explicitness is a necessary condition for progress in evaluating complex socio-cognitive capabilities in language models.

\section{Limitations}
There are some limitations regarding the utility and applicability of TTC that constrain the extent to which a TTC facilitates correct interpretation and generalization of LLM capabilities and subsequent downstream use and deployment. Making theoretical assumptions explicit does not ensure that those assumptions are accurate, nor does it eliminate judgment calls in how a card is completed. Different authors may reasonably fill out the same TTC differently, reflecting genuine theoretical disagreement about how a capability should be defined or decomposed. However, such disagreement in TTC completion is itself informative, surfacing theoretical divergence that is otherwise implicit.
In addition, because the TTC relies on human construct concepts to structure interpretation, it does not by itself prevent anthropomorphic readings of model behavior or determine whether a given construct framework is appropriate for describing machine behavior. The TTC constrains what claims are supported by an evaluation under stated assumptions, but it does not adjudicate between competing theories. As with other documentation practices, the TTC is most effective when used alongside complementary evaluation methods, such as targeted stress tests or audits that probe claims extending beyond a benchmark’s stated scope.

% --- References ---
%\clearpage
\sloppy
\bibliography{custom}
\clearpage
\appendix

\section{Appendix}
\label{sec:appendix}

%%%%%%%%%%%%%%%%%%%%%%%%%%%%%%%%%%%%%%%%%%%%%%%%%%%%%
\begin{tcolorbox}[
  title={Theory Trace Card for EmpatheticDialogues {\hypersetup{citecolor=white}\citep{rashkin2019empathetic}}},
  breakable,
  enhanced,
  width=\textwidth,
  fontupper=\small
]

\textbf{1. Theory}
\begin{itemize}
  \item \textbf{Framework:} Appraisal Theory of Empathy \citep{wondra2015appraisal} \textcolor{red}{[The authors' operational framework is ``Empathetic Response,'' which we map here to the closest theoretical framework for empathy, namely the Appraisal Theory of Empathy.]}
  \item \textbf{Core components:}
  \begin{itemize}
    \item Appraisal of the target's situation 
    \item Vicarious emotional experience 
    \item Compassion
  \end{itemize}
\end{itemize}

\vspace{0.5em}

\textbf{2. Components Exercised}
\begin{itemize}
    \item Appraisal of the target's situation
    \item Compassion
\end{itemize}

\vspace{0.5em}

\textbf{3. Task Operationalization}
\begin{itemize}
  \item \textbf{Task:} The model acts as a ``Listener'' and must generate a response to a ``Speaker'' describing a personal situation
  %Given a single utterance from a speaker describing a personal situation, the model generates a response intended to be empathetic toward the speaker.
  \item \textbf{Key specs:} 
  \begin{itemize} \item Input: Dialogue history (context) grounded in one of 32 emotion labels (e.g., Proud, Afraid). \item Constraints: The model has access to the text context but not to the emotion label itself. \end{itemize}
  
\end{itemize}

% \vspace{0.5em}

 \textbf{Scoring Criterion:}
 \begin{itemize}
   \item Automated similarity-based metrics (e.g., BLEU; \citealp{papineni2002bleu}) and human evaluations assessing the perceived appropriateness and empathy of generated responses relative to reference replies.
 \end{itemize}

\vspace{0.5em}
\textbf{4. Inference and Limitations}
\begin{itemize}
  \item \textbf{Inference:} Performance supports cognitive empathy.
  \item \textbf{Limitations:} No overlapping emotions or non-textual cues (tone, body language). Only tested short scenarios. Dialogues sourced from MTurk workers whose demographics are known to be culturally biased \citep{paolacci2010running}.
\end{itemize}

\end{tcolorbox}

%\clearpage

%%%%%%%%%%%%%%%%%%%%%%%%%%%%%%%%%%%%%%%%%%%%%%%%%%%%%%%%%%%

\begin{tcolorbox}[
  title={Theory Trace Card for LLM ToM Evaluation {\hypersetup{citecolor=white}\citep{kosinski2024evaluating}}},
  breakable,
  enhanced,
  width=\textwidth,
  fontupper=\small
]

\textbf{1. Theory}
\begin{itemize}
  \item \textbf{Framework:} Theory of Mind \citep{wimmer1983beliefs,perner1987three,heyes2014cultural}.
  \item \textbf{Core components:} \textcolor{red}{[\textit{Not explicitly stated in paper}.]}
  \begin{itemize}
      \item Cognitive ToM (e.g., belief-tracking)
      \item Affective ToM (e.g., emotion-tracking)
  \end{itemize}

\end{itemize}

\vspace{0.5em}

\textbf{2. Components Exercised}
\begin{itemize}
    \item Cognitive ToM (belief tracking)
\end{itemize}

\vspace{0.5em}

\textbf{3. Task Operationalization}
\begin{itemize}
  \item \textbf{Task:} Given a short, structured narrative describing an agent, an object, and a belief-relevant change in the environment, the model answers questions predicting the agent’s belief or action.
  \item \textbf{Key specs:} Tasks are modeled after classic developmental false-belief paradigms, including ``Smarties'' and ``Sally–Anne'' tasks \citep{wimmer1983beliefs,perner1987three}. Each false-belief scenario is paired with closely matched true-belief control scenarios and reversed versions to control for task structure and language cues.
  \item \textbf{Scoring Criterion:} Accuracy measured as the proportion of scenarios fully solved, where a scenario is counted as correct only if all sub-questions (false-belief and control questions) are answered correctly. 
\end{itemize}

\vspace{0.5em}

\textbf{4. Inference and Limitations}
\begin{itemize}
  \item \textbf{Inference:} Performance supports cognitive ToM (belief-tracking).
  \item \textbf{Limitations:} Does not cover non-text-based cues (e.g., gaze). Only short text-based scenarios. False-belief tasks were developed for Western populations, and performance may be culturally biased \citep{lillard1998ethnopsychologies,heyes2014cultural}.
  \end{itemize}

\end{tcolorbox}

\clearpage

%%%%%%%%%%%%%%%%%%%%%%%%%%%%%%%%%%%%%%%%%%%%%%%%%%%%

\begin{tcolorbox}[
  title={Theory Trace Card for \textit{ETHICS} Moral Reasoning {\hypersetup{citecolor=white}\citep{hendrycks2021ethics}}},
  breakable,
  enhanced,
  width=\textwidth,
  fontupper=\small
]

\textbf{1. Theory}
\begin{itemize}
  \item \textbf{Framework:} A multi-theory account of moral reasoning drawing on justice \citep{sidgwick1907methods}, deontology \citep{rawls1999collected}, virtue ethics \citep{aristotle340bc}, utilitarianism \citep{de2017utilitarianism}, and commonsense moral judgment \citep{reid1788essays} %\citep{hendrycks2021ethics}.
  \item \textbf{Core components:} 
  \begin{itemize}
   \item Justice. 
   \item Deontological reasoning. 
  \item Virtue and vice attribution. 
   \item Utilitarian reasoning. 
   \item Commonsense moral reasoning.
   \end{itemize}

\end{itemize}

\vspace{0.5em}

\textbf{2. Components Exercised}
\begin{itemize}
  \item Justice.
  \item Deontological reasoning.
  \item Virtue and vice attribution.
  \item Utilitarian reasoning.
  \item Commonsense moral judgment.
\end{itemize}

\vspace{0.5em}

\textbf{3. Task Operationalization}
\begin{itemize}
  \item \textbf{Task:} Given a short, stylized moral scenario, the model produces a discrete judgment aligned with the normative framing of the task (e.g., reasonable vs.\ unreasonable, virtue vs.\ vice, more vs.\ less pleasant).
  \item \textbf{Key specs:} The benchmark comprises multiple task types, one for each of the components above. All tasks use fixed prompts and closed-set response formats.
  \item \textbf{Scoring Criterion:} Accuracy with respect to human-labeled judgments for each task type. For Justice, Deontology, Virtue Ethics, and Commonsense Morality, responses are scored based on agreement with annotated reasonableness or acceptability labels; for Utilitarianism, scoring reflects correct identification of the more pleasant (lower-pain) scenario in paired comparisons.

\end{itemize}

\vspace{0.5em}

\textbf{4. Inference and Limitations}
\begin{itemize}
  \item \textbf{Inference:} Performance supports moral reasoning in unambiguous text-based scenarios in terms of justice, deontology, virtue ethics, utilitarianism, and commonsense moral intuitions.
  \item \textbf{Limitations:} Excludes multimodal, sequential, interactive, and open-ended reasoning. Benchmark operationalizes ethics primarily through Western moral theories (e.g., deontology, utilitarianism, virtue ethics). Primarily includes data from English speakers from the United States, Canada, and the United Kingdom, sourced from MTurk and Reddit. 

\end{itemize}

\end{tcolorbox}

%\clearpage

%%%%%%%%%%%%%%%%%%%%%%%%%%%%%%%%%%%%%%%%%%%%%%%%%%%%

\clearpage

\begin{tcolorbox}[
  title={Theory Trace Card for \textit{GoEmotions} {\hypersetup{citecolor=white}\citep{demszky2020goemotions}}},
  breakable,
  enhanced,
  width=\textwidth,
  fontupper=\small
]

\begin{itemize}
  \item \textbf{Framework:} Emotion Taxonomy after \citealp{cowen2017self}.
  \item \textbf{Core components:} Emotion recognition
  {\color{red}
  [Paper tests ability to recognize the emotion categories identified in Cowen and Keltner, 2017]
  }
\end{itemize}

\vspace{0.5em}

\textbf{2. Components Exercised}
\begin{itemize}
  \item Emotion recognition.
\end{itemize}

\vspace{0.5em}

\textbf{3. Task Operationalization}
\begin{itemize}
  \item \textbf{Task:} Given a single short text comment, the model predicts one or more emotion labels from a predefined set.
  \item \textbf{Key specs:} Single-utterance inputs (Reddit comments); closed-set label space consisting of 27 emotion categories plus neutral; multi-label classification; no justification required.
  \item \textbf{Scoring Criterion:} Performance is evaluated by agreement with human-annotated emotion labels, using standard multi-label classification metrics.
\end{itemize}

\vspace{0.5em}

\textbf{4. Inference and Limitations}
\begin{itemize}
  \item \textbf{Inference:} Performance supports the model's ability to recognize clearly expressed emotions in short text snippets under a fixed category schema
  \item \textbf{Limitations:} Does not evaluate other facets of emotion understanding (e.g., detecting subtle or implicit emotions, understanding causes of emotions). Emotion taxonomy for labeling texts was created based on MTurk ratings without controls for cultural diversity.
\end{itemize}

\end{tcolorbox}

%%%%%%%%%%%%%%%%%%%%%%%%%%%%%%%%%%%%%%%%%%%%%%%%%%%%%

\begin{tcolorbox}[
  title={Theory Trace Card for \textit{SocialIQA} {\hypersetup{citecolor=white}\citep{sap2019socialiqa}}},
  breakable,
  enhanced,
  width=\textwidth,
  fontupper=\small
]

\textbf{1. Theory}
\begin{itemize}
  \item \textbf{Framework:} Commonsense Psychology \citep{moore2013development}
  \item \textbf{Core components:} 
  \begin{itemize}
  \item Inferring motivations. 
  \item Inferring next actions. 
  \item Inferring emotions.
  \end{itemize}

\end{itemize}

\vspace{0.5em}

\textbf{2. Components Exercised}
\begin{itemize}
  \item Inferring motivations.
  \item Inferring next actions.
  \item Inferring emotions.
\end{itemize}

\vspace{0.5em}

\textbf{3. Task Operationalization}
\begin{itemize}
  \item \textbf{Task:} Given a short narrative describing an everyday social situation, the model answers a multiple-choice question about motivations and intentions, emotional reactions, or likely next actions.
  \item \textbf{Key specs:}  Single sentence scenarios; multiple-choice format with three answer options; questions and answers constructed using a combination of crowdsourced annotations and prior datasets.
  \item \textbf{Scoring Criterion:} Accuracy measured as agreement with human-annotated correct answers collected via MTurk.
  % \item \textbf{Scoring inference:} Selecting the human-annotated correct answer is treated as evidence that the model successfully applied social commonsense reasoning to interpret the situation.
\end{itemize}

\vspace{0.5em}
\textbf{4. Inference and Limitations:}
\begin{itemize}
   \item \textbf{Inference:} Performance supports social commonsense reasoning.
    \item \textbf{Limitations:} Uses constrained, textual scenarios with limited context. Scenarios were sourced from English-speaking WEIRD sources and annotated by MTurk workers, which may be culturally biased and not reflect non-Western social norms.
\end{itemize}

\end{tcolorbox}

\end{document}